\documentclass[times, review, 10pt]{elsarticle} 
\usepackage[top=4.3cm, right=4.8cm, bottom=4.3cm, left=4.8cm]{geometry}

\usepackage{amsmath}
\usepackage{amssymb}
\usepackage{caption}
\usepackage{booktabs}
\usepackage{graphicx}
\usepackage{multirow}   

\biboptions{numbers,sort&compress}

\title{DCSNet: Multiscale Feature Aggregation for Small Medical Object Segmentation with Detection-guided Hierarchical Cropping}

\author[label1]{Shanfeng Zhang\fnref{1}}
\author[label1,label3]{Bo Gou\fnref{1}}
\author[label1]{Yue Cao}
\author[label2]{Lei Zhang}
\author[label1]{Zhang Yi}
\author[label2]{Tao He\corref{cor1}}
\ead{tao_he@scu.edu.cn}
\cortext[cor1]{Corresponding author.} 

\fntext[1]{These authors are co-first authors and contributed equally to this work.} 

\affiliation[label1]{organization={College of Computer Science, Sichuan University},
                postcode={610065},
                city={Chengdu},
                country={China}}
\affiliation[label2]{organization={School of Artificial Intelligence, Sichuan University},
                postcode={610065},
                city={Chengdu},
                country={China}}
\affiliation[label3]{organization={The First Affiliated Hospital of Chengdu Medical College, School of Clinical Medicine, Chengdu Medical College},
postcode={610065},
            city={Chengdu},
            country={China}}
\begin{document}
 
\begin{abstract}
    Small object segmentation in medical imaging is primarily hindered by class imbalance and inherent boundary complexity. Consequently, conventional global networks frequently fail to detect sparse targets or suffer from severe edge degradation. To overcome these limitations, we propose the Detection-guided Cropping Segmentation Network (DCSNet), an end-to-end framework that transforms global dense prediction into a localized refinement process. This framework integrates two core components, namely Detection-guided Hierarchical Cropping (DGHC) and Multiscale Feature Aggregation (MSFA). The DGHC module leverages region proposals to dynamically extract object-centric features, effdataectively filtering out massive background interference to mitigate class imbalance. Subsequently, the MSFA module operates strictly within these purified regions, synergizing a Transformer encoder with a pixel-adaptive fusion strategy. This mechanism dynamically aggregates multiscale features to capture both semantic context and fine-grained details for sharp boundary delineation. Extensive experiments across three diverse medical datasets demonstrate that DCSNet significantly outperforms existing state-of-the-art methods, yielding substantial improvements in boundary precision and offering a highly robust solution for clinical micro-lesion segmentation.
\end{abstract}

\maketitle

\section{Introduction}
\begin{figure}[htb]
    \centering
    \includegraphics[width=1.0\linewidth]{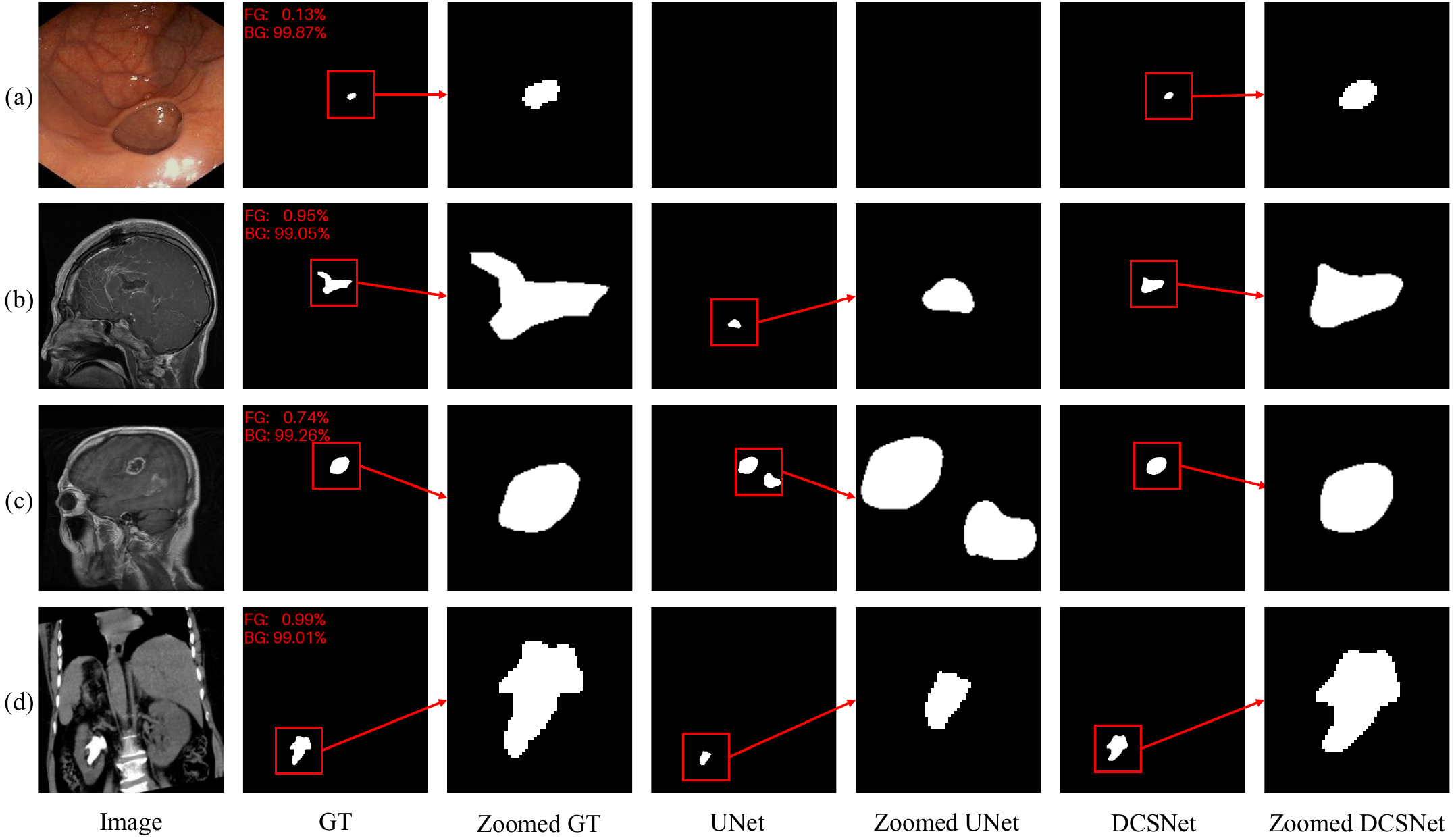}
    \caption{Visual comparison between UNet and DCSNet. All foreground targets occupy less than 1\% of total image pixels. Extreme class imbalance induces segmentation failures (e.g., false negatives) in (a-b). Inherent boundary complexity  causes boundary-level contour distortions (e.g., severe leakage and shrunken edges) in (c-d). In contrast, our method achieves more complete and precise segmentation. }
    \label{fig:visual_comparison_v3}
\end{figure}

    Medical image segmentation, as a core part of Computer-Aided Diagnosis (CAD), plays an irreplaceable role in  disease screening, treatment planning and efficacy monitoring \cite{rodriguez2016computer}. It is widely applied in the analysis of various medical imaging modalities, including CT, MRI, and endoscopic imaging \cite{kumar2022deep}. Small object segmentation specifically refers to a low proportion in the image or small-sized anatomical structures. These small objects are the key evidence for early disease diagnosis, and segmentation accuracy directly decides the clinical application values of CAD systems. However, small object segmentation in medical images faces two primary inherent challenges. First, severe class imbalance. Small targets occupy a minuscule fraction of the overall spatial area compared to the background tissue. Following established criteria \cite{kong2024efcnet}, we define small objects as targets occupying less than 1\% of the total image pixels. This extreme disparity forces models to heavily bias towards the background during optimization, leading to inadequate feature learning for small objects. As shown in Fig.~\ref{fig:visual_comparison_v3} (a) and (b), such optimization bias causes conventional models like UNet \cite{ronneberger2015u} to fail in capturing sparse foreground features, manifesting false negatives. Second, inherent boundary complexity. Small medical objects naturally exhibit irregular shapes, blurred boundaries, and poor separability from surrounding tissues. The inherently low contrast in diverse imaging modalities acts as a direct catalyst that further exacerbates this poor separability \cite{tajbakhsh2020embracing}. Consequently, segmentation models struggle to extract discriminative features and accurately delineate target edges. This deficiency directly translates into various segmentation errors, resulting in severe boundary leakage into surrounding tissues as illustrated in Fig.~\ref{fig:visual_comparison_v3} (c), and edge degradation as illustrated in Fig.~\ref{fig:visual_comparison_v3} (d), where highly irregular jagged contours are severely shrunken.
    
    To tackle the dual challenges of extreme class imbalance and inherent boundary complexity, existing methods have explored numerous strategies but still remain the core bottleneck. Mainstream CNN-based architectures, particularly UNet and its variants \cite{ronneberger2015u, zhou2018unet++, das2024attention}, attempt to preserve shallow details through feature extraction and skip connections. However, due to the localized receptive fields of standard convolutions, the representations of small objects are easily overwhelmed by dominant background features or nearby large tissues, leaving these models highly susceptible to extreme class imbalance. To overcome the localized limitations of CNNs and establish global contextual awareness, Vision Transformers and hybrid architectures \cite{chen2021transunet, cao2022swin} have been subsequently introduced into the field. By leveraging self attention mechanisms \cite{vaswani2017attention}, Transformer-based models can effectively build global semantic correlations. Nevertheless, this global modeling paradigm inherently lacks local inductive biases, severely compromising its ability to capture the fine-grained structural details essential for precise boundary delineation \cite{chen2021transunet, hatamizadeh2022unetr}.
    
\begin{figure}[htbp]
    \centering
    \includegraphics[width=0.7\linewidth]{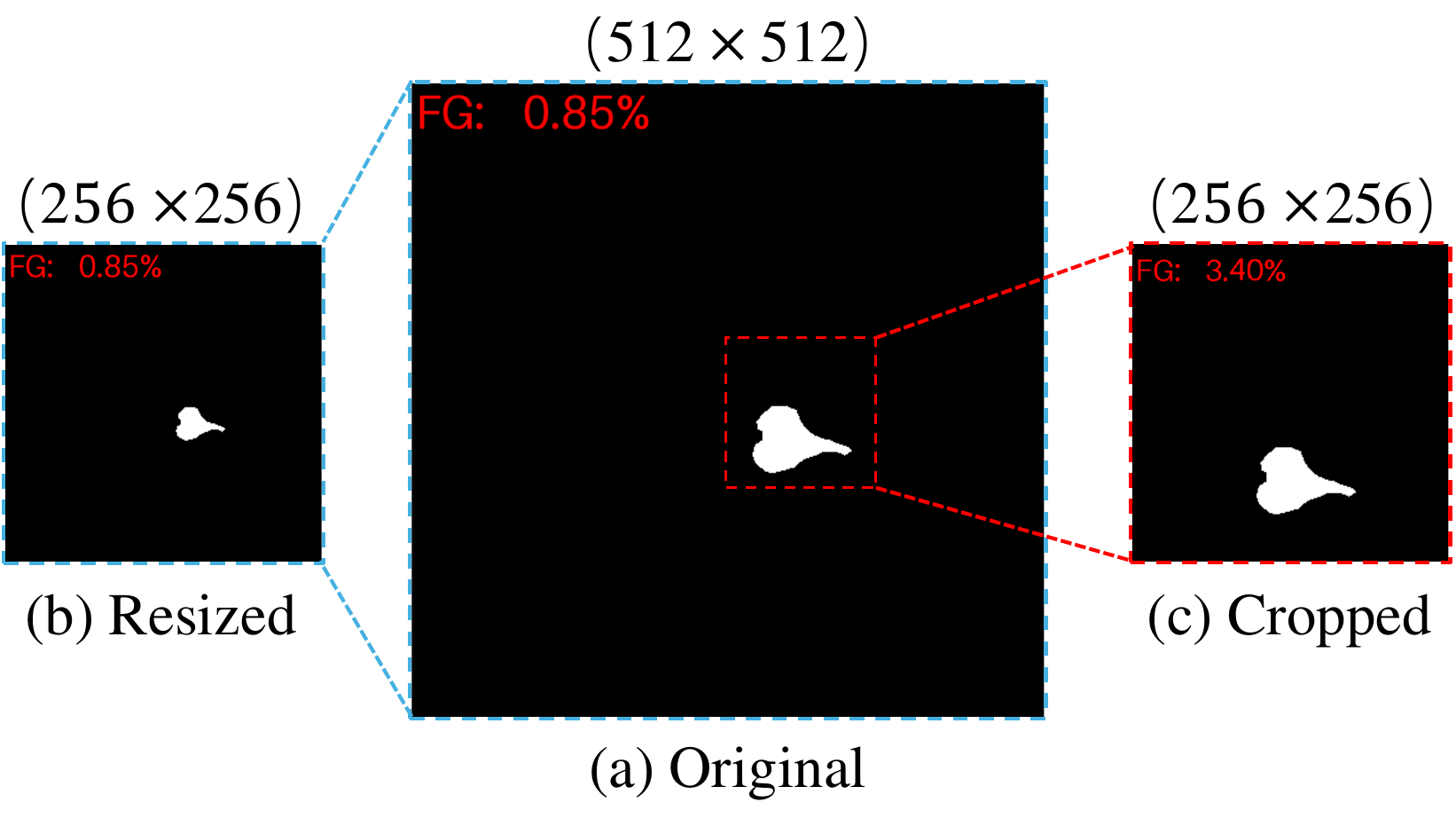}
    \caption{Illustration of the three data-level settings in our preliminary study. (a) Original: global segmentation on original high-resolution images. (b) Resized: global segmentation on directly downsampled images. (c) Cropped: local segmentation utilizing oracle bounding boxes effectively mitigates background interference.}
    \label{fig:experiment_settings}
\end{figure}

    Notably, clinical diagnosis naturally follows a paradigm of localizing the target region prior to detailed morphological analysis. To empirically validate whether such spatial constraints benefit neural networks, we conducted a preliminary study comparing standard global segmentation with strictly localized cropping as illustrated in Fig.~\ref{fig:experiment_settings}. Result in Table~\ref{tab:motivation} demonstrates that explicitly constraining the input to cropped target regions yields significant improvements in both Dice and IoU metrics compared to global paradigms. This substantial performance gain demonstrates that effectively filtering out massive background noise is the primary catalyst for accuracy enhancement, a principle that conceptually aligns with coarse-to-fine cascade strategies in computer vision.

\begin{table}[htbp]
    \centering
    \caption{Quantitative results of the preliminary study on the Brain Tumor dataset. The comparison across different input resolutions demonstrates that explicitly cropping the target region significantly improves performance by mitigating background interference.}
    \label{tab:motivation}
    \begin{tabular}{lccc}
    \toprule
    Methodology Paradigm & Resolution & Dice $\uparrow$ & IoU $\uparrow$ \\
    \midrule
    Global (Original) & $512 \times 512$ & 75.79 & 67.46 \\
    Global (Resized) & $256 \times 256$ & 75.58 & 67.51 \\
    Local (Cropped)  & $256 \times 256$ & \textbf{77.26} & \textbf{69.43} \\
    \bottomrule
    \end{tabular}
\end{table}

    Despite this theoretical alignment, directly deploying decoupled two-stage cascade architectures exposes severe architectural bottlenecks \cite{zhang2021transfuse}. Rigid image-level cropping inherently prevents feature sharing across stages, incurring immense computational redundancy \cite{liu2021review}. More critically, the lack of end-to-end joint optimization inevitably triggers irreversible error accumulation \cite{isensee2021nnu}. Any spatial misalignment during initial localization strictly upper-bounds the subsequent segmentation performance and cannot be dynamically corrected \cite{hatamizadeh2022unetr}. To break these bottlenecks, we propose an end-to-end Detection-guided Cropping Segmentation Network (DCSNet). The main contributions of this work are summarized as follows:
    
    \begin{enumerate}
    \item We propose DCSNet, a novel end-to-end framework tailored for small object segmentation in medical images. Rather than relying on disconnected image-level separation, it seamlessly integrates detection-driven localization with hierarchical feature-level region cropping and subsequent fine-grained segmentation. This fully differentiable architecture transforms an overwhelmingly difficult global task into a highly focused localized refinement problem, effectively bypassing the irreversible error accumulation of conventional two-stage methods.
    
    \item We design a Detection-guided Hierarchical Cropping (DGHC) module to tackle the severe class imbalance. By leveraging region proposals with a scale aware jitter strategy and hierarchical RoIAlign, this module restricts the segmentation attention to object-centric regions, which effectively filters out massive background interference and improves robustness to localization errors.
    
    \item We develop a Multiscale Feature Aggregation (MSFA) module to overcome boundary complexity. This module hierarchically extracts and adaptively fuses multiscale features, enabling the model to simultaneously capture local fine-grained details for sharp boundary delineation and global semantic correlations for structural consistency.

    \end{enumerate}
    
\section{Related Work}
\subsection{Object Detection}
    R-CNN \cite{girshick2014rich} was the first to introduce convolutional neural networks into region based detection. It generates candidate regions through selective search and extracts features, but it has the shortcomings of repetitive computation, slow inference, and module fragmentation \cite{he2015spatial}. To address the efficiency issue, Fast R-CNN \cite{girshick2015fast} adopted feature sharing of the entire image and the Region of Interest (RoI) Pooling technique to achieve end to end training for classification and bounding box regression. However, it still relied on traditional methods to generate candidate regions, making it difficult to optimize the quality of candidate regions. 

    Therefore, Faster R-CNN \cite{ren2015faster} was proposed to introduce the Region Proposal Network (RPN) and replace traditional candidate region generation methods. It shares features with the backbone network to achieve end-to-end two-stage detection, significantly improving detection speed and accuracy. Parallel to this two-stage evolution, one-stage detectors such as YOLOv10 and YOLOv11 architectures \cite{wang2024yolov10, palaniappan2025yolo} have achieved remarkable real time inference in various vision tasks. Despite their computational efficiency, their grid based prediction mechanisms often struggle with the extreme scale imbalance of tiny medical lesions. Consequently, two-stage architectures with dedicated region proposal networks remain the gold standard for high recall localization of micro objects.

    To extend this precise two-stage localization to pixel level prediction, Mask R-CNN \cite{he2017mask} adds a mask branch on top of the Faster R-CNN framework. While it successfully achieves joint detection and segmentation, its mask branch relies on a fixed low resolution spatial output. When applied to medical images with extremely tiny targets, this inherent architectural constraint inevitably leads to over smoothed edges and a loss of fine morphological details. This makes it challenging to perfectly capture highly irregular boundaries in clinical diagnostics \cite{felfeliyan2022improved, kirillov2020pointrend, bozorgpour2023dermosegdiff}.

\subsection{Medical Image Segmentation}
    UNet \cite{ronneberger2015u} has established the fundamental paradigm of medical image segmentation with its symmetric encoder decoder architecture and skip connections. However, its downsampling process can lead to the loss of spatial information \cite{wang2020non}. Subsequent studies have evolved multiple branches to address these structural limitations. UNet++ alleviates the semantic gap between encoders and decoders through nested dense skip connections, improving the fusion accuracy of fine-grained features \cite{zhou2018unet++, huang20203+}. Attention U-Net introduces spatial attention gates to enhance the discriminant power on the region of interest by suppressing background noise \cite{das2024attention, you2024learning}. Building upon these classic paradigms, recent works continue to refine U-like structures by developing advanced multiscale fusion methods \cite{he2025fuseunet, he2024lightweight, cao2026enhancing,  zhang2026high} and strengthening layer interactions via dynamic attention mechanisms \cite{wang2024strengthening, han2026sams}.

    To model global context and capture long range spatial dependencies, Transformers are introduced via self attention operations \cite{xiao2023transformers}. Self supervised paradigms like multiscale Masked Autoencoders further improve representation learning \cite{yu2025bus}. However, their high computational complexity and dependence on large scale data limit their generalization. To address this computational bottleneck, recent cutting edge advancements have decisively shifted towards State Space Models (SSMs) and neural memory ordinary differential equations. Selective State Space Models like Mamba provide global receptive fields with linear computational complexity \cite{kurniawan2025mamba}. Furthermore, continuous ODE based networks like CNM-UNet \cite{niu2024bidirectional, yu2026mobileode, xu2026cnm} offer mathematically elegant continuous spatial dynamics modeling. 

    While these continuous and state space formulations elegantly solve efficiency issues for global context modeling, they fundamentally operate as global dense predictors. Recent studies have demonstrated that without explicit spatial constraints and local priors, the global features extracted by these models often fail to capture fine boundaries and are easily overwhelmed by complex backgrounds in small target scenarios \cite{chattopadhyay2026robustness, atabansi2025dcm}. Specifically, when dealing with highly variable micro lesions, the foreground signal can become negligible during global optimization, causing even advanced architectures like SSMs to suffer from local detail loss and severe missed detections \cite{xu2025mbgnet}. This consensus in recent literature indicates that pure architectural evolution is insufficient for small objects, and explicitly integrating spatial localization and shape guidance strategies remains an indispensable requirement for robust medical image segmentation \cite{lei2023sgu}.

\subsection{Small Object Segmentation}
    As established in previous sections, the extreme categorical imbalance and low spatial resolution inherent to small targets have consistently posed challenges for medical image segmentation. To improve the ability to detect such micro lesions, current research focuses primarily on the introduction of attention mechanisms and the design of two-stage cascade architectures.
    
    Incorporating attention modules is essential for achieving more accurate segmentation of small objects. Its core lies in enhancing model sensitivity to the features of small targets through dynamic weight allocation while suppressing background interference. For example, EFCNet achieves multiscale feature aggregation through cross stage axial attention \cite{kong2024efcnet}. SvANet proposes scale adaptive attention to capture local details across different frequency bands \cite{dai2024svanet}. Recent studies further advance this paradigm by introducing explicit edge aware architectures and boundary prompted gating mechanisms to force the model focus onto ambiguous lesion edges \cite{xia2025eems, fang2025minding}. Despite these remarkable advancements, linear or shallow nonlinear fusion mechanisms still encounter challenges in fully decoupling the overlap between target and background in clinical scenarios with low contrast and low signal to noise ratios. This ongoing difficulty can sometimes lead to boundary blurring \cite{lei2025condseg, urrea2025advances}.

    The cascade strategy progressing from coarse to fine has evolved into a crucial method for handling small targets. This two-stage framework typically involves first locating the approximate region of the target and then performing fine segmentation on the local area. For instance, shape prior based networks crop RoIs based on initial coarse segmentation outputs and then fuse multiscale spatial pyramid pooling with anatomical priors \cite{wu2023coarse}. CaraNet employs a local encoder for precise localization to enhance the detailed representation of small organs \cite{lou2022caranet}. To further address boundary degradation, G-CASCADE integrates multiscale contextual features within its cascade architecture to capture global and local boundary correlations simultaneously \cite{rahman2024g}. While these cascade frameworks demonstrate excellent refinement capabilities, recent analyses indicate that their overall performance is inherently tied to the accuracy of the initial segmentation stage. In cases where the initial mask based localization is missed or exhibits deviations, the subsequent refinement process faces considerable obstacles due to severe error propagation from the imperfect coarse inputs \cite{mehta2021propagating}. Because these frameworks fundamentally rely on coarse segmentation masks to generate RoIs, they can be susceptible to redundant background inclusion or incomplete target cropping \cite{wang2026hierarchical}. Recognizing this architectural dependency highlights the potential benefits of shifting towards robust detection driven and bounding box guided cropping paradigms \cite{jaeger2020retina, ndir2025dynamic}.
    
\section{Methodology}
\begin{figure}[htb]
    \centering
    \includegraphics[width=1.0\linewidth]{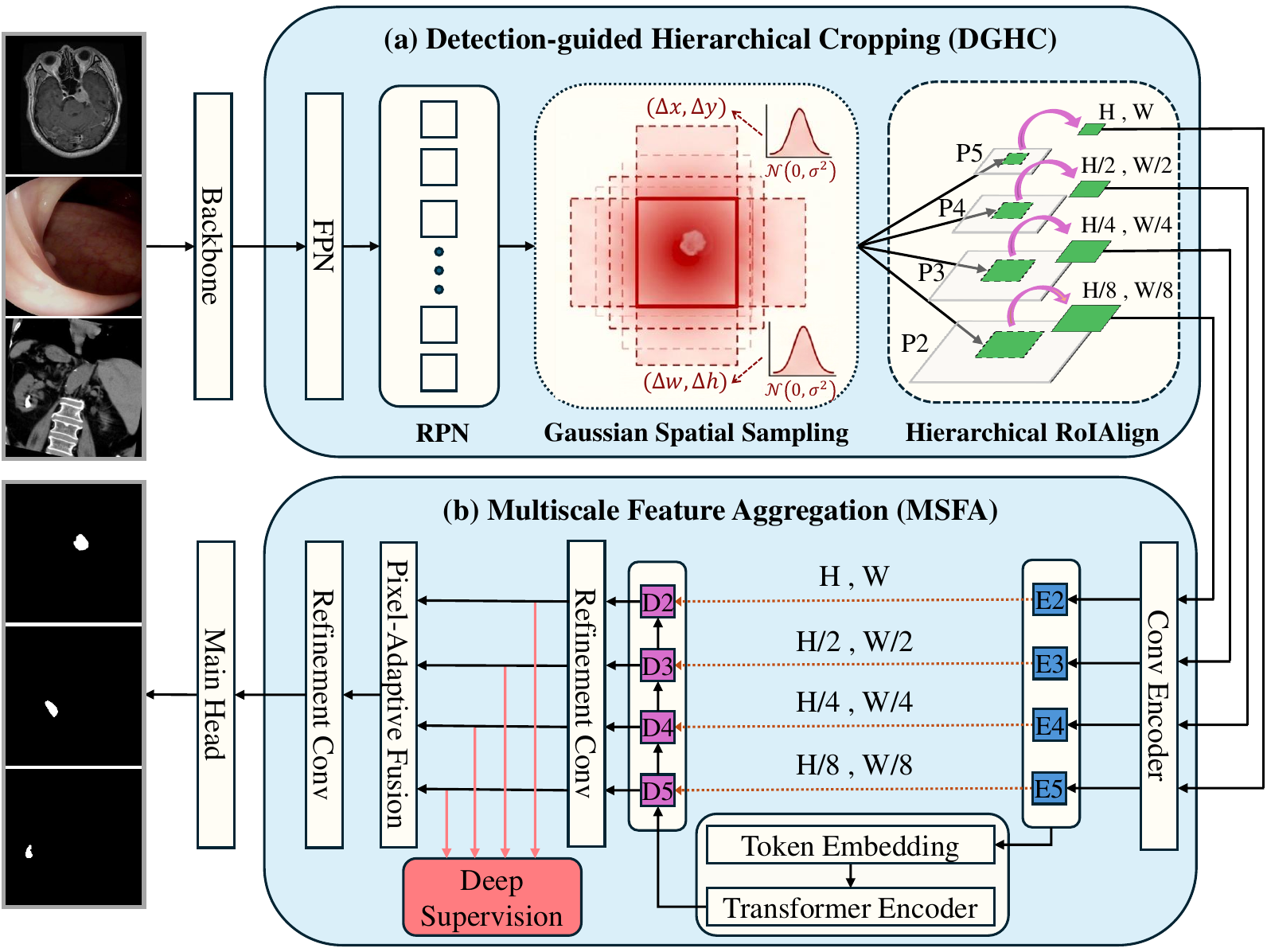}
    \caption{Overview of the proposed Detection-guided Cropping Segmentation Network (DCSNet). (a) Details of the Detection-guided Hierarchical Cropping (DGHC) module, where deterministic region proposals generated by RPN are recast via a Gaussian spatial sampling mechanism to model localization uncertainty, followed by hierarchical RoIAlign on FPN features. (b) Details of the Multiscale Feature Aggregation (MSFA) module, which encodes and decodes cropped features across multiple scales, establishes global dependencies via a Transformer encoder, and produces the final segmentation through pixel-adaptive fusion.}
    \label{fig:framework}
\end{figure}

\subsection{Overall architecture}
    To address the dual challenges of extreme class imbalance and inherent boundary complexity in small object segmentation in medical images, we propose an end-to-end Detection-guided Cropping Segmentation Network (DCSNet). As illustrated in Fig.~\ref{fig:framework}, DCSNet comprises the Detection-guided Hierarchical Cropping (DGHC) module and the Multiscale Feature Aggregation (MSFA) module, which together accomplish the transition from global dense prediction to localized refinement. Given an input medical image, the model first utilizes a Faster R-CNN backbone to construct a multilevel Feature Pyramid Network (FPN) and generates candidate region proposals. Rather than passively accepting these deterministic bounding boxes, the DGHC module introduces a Gaussian spatial sampling mechanism to explicitly model localization uncertainty and mitigate the risk of spatial misalignment. Based on these probabilistically perturbed regions, the module performs hierarchical RoIAlign directly on the FPN features. This uncertainty-aware, object-centric cropping mechanism dynamically extracts fixed-size hierarchical representations, which substantially filters out massive background interference while preserving both deep semantics and shallow structural contexts.

    Subsequently, the cropped hierarchical features are fed into a hybrid segmentation branch designed to integrate local details with global dependencies. To overcome the limited receptive field of standard convolutions, the multilevel features are projected into unified tokens equipped with positional and level embeddings, and they are then processed by a Transformer encoder to guarantee global semantic consistency across all scales. During the decoding stage, the MSFA module abandons rigid linear addition in favor of a spatially aware integration strategy. It adaptively merges the decoded multilevel features using a pixel-wise weighting mechanism, allowing the model to dynamically select the most relevant scale for each specific pixel. Ultimately, this pixel-adaptive fusion enables DCSNet to generate highly accurate and boundary-crisp segmentation masks, thereby bridging the gap between rough target localization and fine-grained boundary delineation.

\subsection{Detection-guided Hierarchical Cropping Module}
    Conventional medical image segmentation networks typically perform dense predictions over the entire global image. For small objects that occupy merely a fraction of the total pixels, this paradigm results in a severe foreground-to-background imbalance, causing models to overwhelmingly bias towards the background and fail to learn discriminative target features. While recent two-stage frameworks attempt to alleviate this imbalance via RoI cropping \cite{wu2023coarse, lou2022caranet}, such paradigms impose a rigid architectural dependency on initial localization precision. The inherently limited spatial footprints and faint semantic signals of small targets render their initial bounding boxes highly susceptible to spatial misalignment. Under conventional rigid cropping mechanisms, even marginal localization offsets inevitably precipitate target truncation or the loss of essential surrounding context, thereby strictly upper-bounding subsequent segmentation performance \cite{lei2023sgu, zhu20183d}.

    To break this bottleneck, we propose the DGHC module. Instead of passively accepting potentially flawed detection proposals, DGHC transforms global segmentation into a localized, scale aware, and uncertainty resilient refinement process. Specifically, we explicitly model the spatial uncertainty of detection boxes during training by introducing a Gaussian spatial sampling mechanism. By applying probabilistic spatial perturbations derived from a Gaussian prior, we force the model to learn robust invariant representations even under severe localization noise. Furthermore, rather than extracting features from a single scale, we employ a hierarchical RoIAlign mechanism. This ensures that the cropped object-centric regions simultaneously capture the fine-grained structural details of the small target and the macro-level semantic context of the surrounding tissues.

    Given multiscale feature maps $\{P_2, P_3, P_4, P_5\}$, a Region Proposal Network (RPN) generates a set of $N$ candidate bounding boxes, denoted as $\mathcal{B}$:
    \begin{flalign}
        \mathcal{B}=\{b_i\}_{i=1}^N, \quad b_i=(x_1,y_1,x_2,y_2) &&
    \end{flalign}
    where $b_i \in \mathcal{B}$ represents the $i$-th proposal, and $(x_1, y_1)$ and $(x_2, y_2)$ denote its top-left and bottom-right spatial coordinates, respectively.

    Directly relying on these proposals renders the segmentation model highly vulnerable to localization errors. This bottleneck is severely amplified for small medical objects, where limited spatial footprints and extreme class imbalance naturally produce noisy bounding boxes. To resolve this, we formulate the deterministic cropping as a stochastic spatial uncertainty modeling process under a Bayesian perspective.

    Specifically, each proposal $b_i$ is first reparameterized into its center coordinates $(x_c, y_c)$ and spatial dimensions $(w, h)$, where $x_c = \frac{x_1+x_2}{2}$ and $w = x_2-x_1$ (with $y_c$ and $h$ derived analogously). Instead of being treated as a definitive bound, these geometric parameters are considered as statistical expectations to construct a probabilistically perturbed bounding box ${\widetilde{b}}_i$ with center $({\widetilde{x}}_c, {\widetilde{y}}_c)$ and dimensions $({\widetilde{w}}, {\widetilde{h}})$:
    \begin{flalign}
        {\widetilde{x}}_c = x_c + \Delta x, \quad {\widetilde{y}}_c = y_c + \Delta y \\
        {\widetilde{w}} = w + \Delta w, \quad {\widetilde{h}} = h + \Delta h
        \end{flalign}
    where the spatial offsets $(\Delta x, \Delta y)$ and scale offsets $(\Delta w, \Delta h)$ are independently sampled from scale-adaptive Gaussian prior distributions:
    \begin{flalign}
        \Delta x \sim \mathcal{N}(0, (\sigma_c w)^2), \quad \Delta y \sim \mathcal{N}(0, (\sigma_c h)^2) \\
        \Delta w \sim \mathcal{N}(0, (\sigma_s w)^2), \quad \Delta h \sim \mathcal{N}(0, (\sigma_s h)^2)
    \end{flalign}
    with $\mathcal{N}(\mu, \sigma^2)$ denoting the normal distribution. The hyperparameters $\sigma_c$ and $\sigma_s$ control the standard deviations of the center shift and scale variance, respectively, both set to 0.05. To prevent out-of-bounds spatial sampling, all perturbed bounding box coordinates are directly clamped within the absolute image boundaries.

    This formulation explicitly models the spatial uncertainty of bounding box localization. By injecting Gaussian-distributed spatial noise, the model is encouraged to learn shift-invariant representations under spatial misalignment. Moreover, unlike uniform distributions, Gaussian sampling natively reflects the physical reality of medical lesions, where the core probability is highest and gradually decays towards the uncertain periphery. This probabilistically expanded sampling region implicitly incorporates essential surrounding contextual cues, thereby effectively preventing target truncation and compensating for initial localization offsets.To ensure stable and reproducible predictions during inference, this stochastic perturbation is strictly disabled, and the model relies solely on the deterministic RPN proposals.

    Based on these probabilistically perturbed regions, we perform feature-level hierarchical cropping using RoIAlign:
    \begin{flalign}
        F_l^{\left(i\right)}=RoIAlign(P_l,{\widetilde{b}}_i), \quad l\in\{2,3,4,5\} &&
    \end{flalign}
    where $F_l^{\left(i\right)}$ denotes the localized feature representation extracted from the $l$-th level $P_l$ of the FPN for the $i$-th perturbed proposal ${\widetilde{b}}_i$. To maintain a standardized input for the subsequent convolutional encoder, the RoIAlign operation directly pools these hierarchical features into a fixed-resolution pyramid of $H \times W$, $\frac{H}{2} \times \frac{W}{2}$, $\frac{H}{4} \times \frac{W}{4}$, and $\frac{H}{8} \times \frac{W}{8}$ corresponding to levels $P_2$ through $P_5$, respectively. By defining the base scale as $H=W=64$ in our implementation, this hierarchical structure ensures that the segmentation head consistently processes object-centric features regardless of the original proposal size.
    
    This hierarchical design enables complementary multiscale perception within a unified RoI. High-resolution features preserve local structural integrity, while low-resolution features encode macro-level contextual cues. Aligning these diverse representations directly facilitates joint reasoning over local geometry and global semantics. Ultimately, by transforming global dense prediction into an uncertainty-resilient, object-centric cropping mechanism, DGHC effectively neutralizes extreme class imbalance and localization noise. The resulting standardized, multiscale feature pyramid provides a robust architectural foundation for the subsequent pixel-adaptive fusion stage.

\subsection{Multiscale Feature Aggregation Module} 
    While the DGHC module effectively localizes targets and alleviates extreme class imbalance, delineating precise contours remains a critical challenge due to inherent boundary complexity. In standard encoder-decoder architectures, multiscale features are typically integrated via linear operations such as concatenation or element-wise addition \cite{ronneberger2015u, zhou2018unet++}. However, these spatially rigid fusion mechanisms inherently struggle with the severe edge degradation and poor separability typical of small medical objects \cite{tajbakhsh2020embracing}. A uniform fusion strategy inevitably compromises local structural details, precipitating semantic ambiguity and over-smoothed edges.
    
    To overcome this limitation, we develop the MSFA module. Rather than adopting conventional plug-and-play fusion components, MSFA is explicitly tailored for the highly dynamic, region-specific features extracted by DGHC. It seamlessly couples global dependency modeling with pixel-wise scale adaptation. Specifically, a Transformer encoder is deployed to bypass the restricted receptive fields of local convolutions, establishing long-range contextual dependencies for global structural consistency \cite{chen2021transunet}. To subsequently counteract the boundary over-smoothing inherent to such global modeling \cite{cao2022swin}, a pixel-adaptive fusion mechanism is introduced to dynamically compute spatially varying weights, empowering DCSNet to select the optimal feature scale at the pixel level.

    Specifically, let $\{E_2, E_3, E_4, E_5\}$ denote the hierarchical features extracted by the convolutional encoder. To construct a unified representation space, these spatial features are transformed and assembled. By incorporating learnable positional embeddings ($E_{pos}$) and level embeddings ($E_{lvl}$), the global multiscale token sequence $T$ is formulated as:
    \begin{flalign}
        T=Concat(Tokens(E_2,E_3,E_4,E_5))+E_{pos}+E_{lvl} &&
    \end{flalign}
    where $Tokens(\cdot)$ denotes the sequential operations of spatial flattening and linear projection applied to each feature map $E_i$, and $Concat(\cdot)$ represents the concatenation of the resulting tokens along the sequence dimension. This sequence is subsequently processed by a multilayer Transformer encoder. Unlike standard Vision Transformers that operate globally across the entire image and inevitably suffer from severe attention dilution in small-target scenarios, the self attention mechanism herein is strictly confined within the object-centric cropped regions. This architectural constraint effectively prevents representational collapse, allocating the full attention capacity to modeling the intricate structural relationships between the lesion and its immediate surrounding tissue.
    
    Building upon this global contextual foundation, the token sequence is reshaped and progressively upsampled through a decoder hierarchy to reconstruct the multilevel spatial features $\{D_2, D_3, D_4, D_5\}$. These intermediate representations first undergo a lightweight convolutional refinement to enhance local spatial details and stabilize the representation space. At this stage, these refined levels serve a dual architectural purpose by providing auxiliary gradient signals to facilitate the stable optimization of DCSNet via deep supervision, while simultaneously acting as the baseline inputs for the subsequent feature integration. Despite their rich semantics, directly aggregating these hierarchical features across divergent scales inevitably introduces spatial mismatches. To alleviate the boundary ambiguity caused by such disparities, a pixel-adaptive fusion mechanism is deployed. Spatially varying weights $W$ are dynamically computed via a dedicated gating operation:
    \begin{flalign}
        W=Softmax(g(Concat(D_2, D_3, D_4, D_5))) &&
    \end{flalign}
    where $g(\cdot)$ denotes consecutive convolution operations applied to the concatenated multiscale features. The adaptively fused feature $F_{fusion}$ is then obtained by aggregating the scales at a pixel level:
    \begin{flalign}
        F_{fusion}=\sum_{i=2}^{5}W_i\odot D_i &&
    \end{flalign}
    where $W_i$ and $D_i$ represent the normalized spatial weight map and the decoded feature representation at the $i$-th scale, respectively, and $\odot$ denotes element-wise multiplication. Rather than relying on hardcoded rules, this formulation drives the gating convolutions to adaptively learn pixel-level scale preferences during optimization. It empowers DCSNet to autonomously assign higher weights to shallow features at object boundaries and deeper features at the object core. Such data-driven allocation intrinsically preserves morphological integrity and effectively mitigates the edge degradation typically observed in small object segmentation.

    Following this adaptive aggregation, the unified representation $F_{fusion}$ undergoes a concluding convolutional refinement to further harmonize the multiscale contextual cues. This stabilized feature space is then directly projected through the main prediction head to generate the final segmentation mask. 
    
    By replacing conventional rigid skip connections with this spatially-aware integration strategy, the MSFA module effectively reconciles the architectural tradeoff between global semantic consistency and local contour precision. This tailored design offers a reliable and adaptive solution for delineating small medical targets characterized by extreme morphological complexity.

\subsection{Loss Function}
    While the DGHC module mitigates macroscopic class disparity, pixel-wise hard sample optimization and boundary delineation require targeted penalties. The primary segmentation objective ($\mathcal{L}_{seg}$) addresses these localized challenges via a weighted combination of Dice, Focal, and boundary-aware losses, jointly enforcing region consistency and contour precision. This structural supervision is complemented by an auxiliary Binary Cross-Entropy loss ($\mathcal{L}_{edge}$) computed against Laplacian-extracted ground truths to explicitly refine local edges.

    To stabilize multiscale learning and enforce structural continuity across the decoder hierarchy, a consistency regularization term computes the squared $L_2$ norm between adjacent activated intermediate predictions:
    \begin{flalign}
        \mathcal{L}_{cons}=\frac{1}{L-1}\sum_{i=1}^{L-1}{\parallel\sigma(Y_i)-\sigma(Y_{i+1})\parallel}_2^2 &&
    \end{flalign}
    where $L$ denotes the total number of intermediate decoding stages, $Y_i$ represents the raw prediction logits at the $i$-th stage, and $\sigma(\cdot)$ is the sigmoid activation function used to project the logits into a normalized probability space prior to distance computation.
    
    To integrate these diverse optimization goals, the overall training objective explicitly aggregates the primary prediction, the intermediate multiscale outputs, and the structural constraints into a unified deep supervision framework:
    \begin{flalign}
        \mathcal{L}_{total} = \mathcal{L}_{seg}^{main} + \sum_{j=2}^{5} w_j \mathcal{L}_{seg}^{(j)} + \lambda_1\mathcal{L}_{edge} + \lambda_2\mathcal{L}_{cons} &&
    \end{flalign}
    where $\mathcal{L}_{seg}^{main}$ denotes the hybrid segmentation loss derived from the main prediction head, and $\mathcal{L}_{seg}^{(j)}$ represents the auxiliary segmentation loss computed via the deep supervision branch at the $j$-th decoder stage. To prioritize global structure learning at deeper semantic levels, the scale-aware weights $w_j$ are progressively increased for lower-resolution outputs. The overall formulation is governed by the static hyperparameters $\lambda_1$ and $\lambda_2$, which balance the respective contributions of the auxiliary edge ($\mathcal{L}_{edge}$) and consistency ($\mathcal{L}_{cons}$) objectives.

\section{Experiments and results}

\subsection{Datasets}
    We conduct experiments on three publicly available medical imaging datasets: brain tumor MRI, colorectal polyp endoscopy, and kidney stone CT. These datasets cover diverse imaging modalities, providing a comprehensive benchmark for small target segmentation. To accommodate the specific characteristics of each modality, all images and their corresponding ground-truth masks are resized to standardized spatial resolutions. Furthermore, to rigorously evaluate model performance, each dataset is randomly partitioned into training, validation, and testing sets following a 7:1:2 split ratio.

\textbf{Brain Tumor Dataset} \cite{cheng2016retrieval} 
    comprises T1-weighted contrast-enhanced MRI slices. All images are processed at a resolution of $512 \times 512$ to preserve fine-grained structural details. To maintain consistency with our definition of small objects, we filter the dataset to retain only the slices where the annotated tumor occupies less than 1\% of the total image area. This filtering process yields a highly challenging subset of 1,287 images. Based on our partition protocol, the dataset is split into 900 samples for training, 129 for validation, and 258 for testing.

\textbf{Colorectal Polyp Dataset} \cite{jha2019kvasir}  
    is derived from the Kvasir-SEG benchmark for gastrointestinal endoscopy. All images are standardized to a resolution of $256 \times 256$. Similarly, we apply this 1\% area threshold to filter the original dataset, resulting in a subset of 1,443 images. This step ensures that our evaluation focuses specifically on delineating small, early-stage polyps. The dataset is divided into 1,010 images for training, 144 for validation, and 289 for testing.

\textbf{Kidney Stone Dataset} \cite{kssd2023}  
    contains 957 coronal CT images with expert-annotated stone masks. All samples are resized to a resolution of $256 \times 256$. Because kidney stones inherently manifest as micro-targets and naturally satisfy our small object segmentation criterion, we utilize the entire dataset without additional filtering. We use 669 images for training, 96 for validation, and 192 for testing.

\subsection{Implementation Details}
    The proposed DCSNet is implemented in PyTorch and trained on a single NVIDIA GPU. The model is optimized using the AdamW optimizer with an initial learning rate of $2 \times 10^{-4}$ and a weight decay of $1 \times 10^{-4}$. A \textit{ReduceLROnPlateau} scheduling strategy is employed, reducing the learning rate by a factor of 0.5 if the validation loss plateaus for 5 epochs. The batch size is set to 4. To accelerate training and reduce memory footprint, Automatic Mixed Precision is utilized.

\textbf{Input and Preprocessing.}
    To accommodate the varying native resolutions and characteristics of the datasets, we adopt dataset-specific input resolutions during training and testing. Specifically, the images in the brain tumor dataset are resized to $512 \times 512$, while the images in the colorectal polyp and kidney stone datasets are resized to $256 \times 256$. 

\textbf{Data Augmentation.}
    Given the inherently limited scale of medical imaging datasets, a comprehensive data augmentation strategy is deployed during training to mitigate overfitting and enhance the representational robustness of small lesions. This strategy incorporates both spatial transformations (e.g., random horizontal/vertical flipping, shift-scale-rotate, and grid distortion) and pixel-level perturbations (e.g., random brightness/contrast, gamma adjustment, Gaussian noise, and Gaussian blur). 
    
    During inference, to further stabilize boundary predictions against extreme morphological variations, Test-Time Augmentation is specifically applied to the colorectal polyp and kidney stone datasets. The final segmentation masks for these modalities are derived by ensembling the predictions from multiple augmented inputs. Conversely, standard direct inference without augmentation is retained for the brain tumor dataset and all validation phases.

\textbf{Loss Function and Training Strategy.}
    The loss weights for the main segmentation head are set to $\alpha=0.5$ for Dice loss, $\beta=0.3$ for Focal loss, and $\gamma=0.2$ for boundary loss. The auxiliary loss coefficients are defined as $\lambda_1=0.2$ (edge loss) and $\lambda_2=0.05$ (consistency loss). Furthermore, we dynamically adjust the loss weights of different feature levels across training epochs to stabilize the multiscale learning process. All models are trained for a maximum of 100 epochs with a gradient clipping maximum norm of 1.0. Early stopping is adopted with a patience of 20 epochs based on the total validation loss.

\textbf{Fair Comparison.}
    To ensure fair comparison, all baseline models are trained under the identical hardware environment, data split, preprocessing pipeline, and data augmentation strategy. No additional training data or external pretraining weights are used unless explicitly specified. 

\subsection{Evaluation Metrics}
    To comprehensively evaluate the segmentation performance of the proposed method, we employ four widely adopted metrics: Dice Similarity Coefficient (Dice), Intersection over Union (IoU), Precision, and 95\% Hausdorff Distance (95HD).

    \textbf{Quantitative Results Analysis.}
    Let $TP$, $FP$, $TN$, and $FN$ denote the true positives, false positives, true negatives, and false negatives of the pixel-level predictions, respectively. The regional overlap is measured by Dice and IoU, which are defined as:
    \begin{flalign}
        \operatorname{Dice} = \frac{2 \times TP}{2 \times TP + FP + FN} &&
    \end{flalign}
    \begin{flalign}
        \operatorname{IoU} = \frac{TP}{TP + FP + FN} &&
    \end{flalign}

    The precision reflects the capability of the model to suppress false positives. This is particularly crucial for small object segmentation, where background noise and artifacts can easily be misclassified as valid foreground lesions:
    \begin{flalign}
        \operatorname{Precision} = \frac{TP}{TP + FP} &&
    \end{flalign}
    
    While Dice, IoU, and Precision focus on internal regional overlap, they are less sensitive to boundary quality. Therefore, we utilize 95HD to evaluate the boundary delineation accuracy. HD calculates the maximum distance between the predicted boundary $X$ and the ground truth boundary $Y$. To eliminate the impact of extreme outliers, the 95th percentile of the distances is adopted:
    \begin{flalign}
        95\operatorname{HD}(X, Y) = \max \left( d_{95}(X, Y), d_{95}(Y, X) \right) &&
    \end{flalign}
    where $d_{95}(X, Y)$ denotes the 95th percentile of the directed distances from all points in $X$ to their nearest points in $Y$. A lower 95HD value indicates a higher degree of boundary agreement.

\begin{table}[htbp]
\centering
\caption{Quantitative results on the Brain Tumor test set comparing the proposed method with state-of-the-art methods. The best results are highlighted in bold.}
\label{tab:brain_tumor}
\resizebox{\textwidth}{!}{%
\begin{tabular}{llcccccc}
\toprule
Category & Model & Dice $\uparrow$ & IoU $\uparrow$ & Precision $\uparrow$ & 95HD $\downarrow$ & Params & GFLOPs \\
\midrule
\multirow{4}{*}{Non-pretrained} 
& UNet           & 75.18 & 67.31 & 83.68 & 28.18 & 31.03M & 218.04 \\
& UNet++         & 77.43 & 68.46 & 82.36 & 21.97 &  9.16M & 139.60 \\
& Attention-UNet & 78.05 & 69.72 & 84.84 & 26.65 & 34.88M & 266.53 \\
& CaraNet        & 82.25 & 74.70 & 83.47 & 15.18 & 46.02M &  46.21 \\
\midrule
\multirow{6}{*}{Pretrained}     
& Mask R-CNN      & 79.04 & 70.52 & 80.58 & 15.43 & 43.92M & 262.24 \\
& SvANet         & 78.44 & 70.02 & 83.55 & 21.92 & 177.19M& 118.97 \\
& TransUNet      & 82.20 & 73.75 & 84.68 & 12.67 & 105.91M& 129.26 \\
& TransFuse      & 80.42 & 71.55 & 84.01 & 14.11 & 143.54M&  62.10 \\
& G-CASCADE      & 83.07 & 73.75 & 83.56 & 13.47 & 95.90M &  22.16 \\
& \textbf{DCSNet(Ours)}  & \textbf{83.67} & \textbf{74.04} & \textbf{85.24} & \textbf{12.18} & 59.42M &  83.32 \\
\bottomrule
\end{tabular}%
} 
\end{table}

\subsection{Comparison with State-of-the-Art Methods}
    To demonstrate the effectiveness of the proposed DCSNet, we compare it against various state-of-the-art (SOTA) medical image segmentation models, including classical CNN-based architectures (UNet, UNet++, Attention-UNet), detection-segmentation hybrid models (Mask R-CNN), and recent transformer-based or hybrid networks (CaraNet, SvANet, TransUNet, TransFuse, G-CASCADE). 

\begin{table}[htbp]
\centering
\caption{Quantitative results on the Polyp test set comparing the proposed method with state-of-the-art methods. The best results are highlighted in bold.}
\label{tab:polyp}
\resizebox{\textwidth}{!}{%
\begin{tabular}{llcccccc}
\toprule
Category & Model & Dice $\uparrow$ & IoU $\uparrow$ & Precision $\uparrow$ & 95HD $\downarrow$ & Params & GFLOPs \\
\midrule
\multirow{4}{*}{Non-pretrained} 
& UNet           & 75.44 & 68.76 & 89.35 & 19.74 & 31.04M &  54.51 \\
& UNet++         & 77.94 & 71.99 & 90.48 & 16.70 &  9.16M &  34.91 \\
& Attention-UNet & 83.65 & 73.07 & 85.37 & 12.34 & 34.88M &  66.63 \\
& CaraNet        & 91.71 & 86.44 & 93.07 &  3.51 & 46.64M &  11.50 \\
\midrule
\multirow{6}{*}{Pretrained}     
& Mask R-CNN      & 83.47 & 76.33 & 87.17 &  7.93 & 43.92M &  61.91 \\
& SvANet         & 78.69 & 71.97 & 89.74 & 14.84 & 177.19M&  29.74 \\
& TransUNet      & 91.41 & 85.86 & 93.47 &  3.65 & 105.91M& 32.23 \\
& TransFuse      & 88.96 & 83.71 & 93.89 &  6.08 & 143.54M&  62.10 \\
& G-CASCADE      & 91.75 & 86.73 & 93.52 &  \textbf{3.32} & 30.96M &  5.54 \\
& \textbf{DCSNet(Ours)}  & \textbf{92.97} & \textbf{86.97} & \textbf{94.03} & 5.01 & 59.42M &  46.86 \\
\bottomrule
\end{tabular}%
} 
\end{table}
    
\textbf{Quantitative Results.}
    As shown in Table \ref{tab:brain_tumor}, the proposed DCSNet achieves state-of-the-art performance on the Brain Tumor dataset. The framework obtains the highest Dice score of 83.67\% and an IoU of 74.04\%, outperforming the recent transformer-based G-CASCADE and the classical UNet by margins of 0.6\% and 8.49\% in Dice, respectively. Beyond regional overlap, DCSNet sets a new benchmark in boundary accuracy by achieving the lowest 95HD of 12.18. Since this metric is highly sensitive to contour deviations, such a significant reduction confirms the superior capability of the overall framework in preserving the precise edges of small tumors.

    Polyps typically exhibit severe variations in scale, shape, and contrast against the mucosal background. Table \ref{tab:polyp} illustrates the superiority of DCSNet in handling these complex morphologies. The proposed method delivers a leading Dice score of 92.97\% and an IoU of 86.97\%, establishing substantial improvements in regional overlap over all comparative baselines. While G-CASCADE achieves a marginally lower 95HD (3.32), DCSNet maintains highly competitive boundary precision (95HD of 5.01) alongside its superior global segmentation accuracy. This structural balance ensures robust detection of early-stage, small-scale polyps without sacrificing semantic integrity.

    Kidney stones present unique challenges due to their micro-target nature and inherently low contrast in CT imaging. Table 4 illustrates that DCSNet effectively overcomes these obstacles, achieving a leading Dice score of 77.43\% and the lowest 95HD of 9.90. This performance completely surpasses recent heavyweights like TransUNet (Dice 73.36\%) and G-CASCADE (Dice 72.88\%). Although the classical UNet secures a marginally higher precision (77.69\%), DCSNet maintains a highly competitive precision of 76.87\% while delivering vastly superior boundary delineation (95HD of 9.90 vs. 24.89). Compared to the strong TransUNet baseline, which requires 105.91M parameters, our framework achieves better overall segmentation with nearly half the parameter burden (59.42M). This structural efficiency allows DCSNet to strike an optimal balance, offering state-of-the-art robustness for micro-lesions within a footprint highly suitable for clinical deployment.

\begin{table}[htbp]
\centering
\caption{Quantitative results on the Kidney Stone test set comparing the proposed method with state-of-the-art methods. The best results are highlighted in bold.}
\label{tab:kidney_stone}
\resizebox{\textwidth}{!}{%
\begin{tabular}{llcccccc}
\toprule
Category & Model & Dice $\uparrow$ & IoU $\uparrow$ & Precision $\uparrow$ & 95HD $\downarrow$ & Params & GFLOPs \\
\midrule
\multirow{4}{*}{Non-pretrained} 
& UNet           & 67.39 & 57.87 & \textbf{77.69} & 24.89 & 31.03M &  54.73 \\
& UNet++         & 71.10 & 61.85 & 77.50 & 21.10 &  9.16M &  34.91 \\
& Attention-UNet & 66.33 & 56.93 & 77.23 & 23.90 & 34.88M &  66.63 \\
& CaraNet        & 63.94 & 53.35 & 72.30 & 15.01 & 46.02M &  11.50 \\
\midrule
\multirow{6}{*}{Pretrained}     
& Mask R-CNN      & 69.52 & 57.72 & 75.30 & 16.63 & 43.92M &  25.90 \\
& SvANet         & 67.47 & 57.36 & 74.16 & 18.62 & 177.19M&  29.74 \\
& TransUNet      & 73.36 & 64.02 & 76.47 & 10.06 & 105.91M& 32.23 \\
& TransFuse      & 71.71 & 61.07 & 73.73 & 11.81 & 143.54M&  62.10 \\
& G-CASCADE      & 72.88 & 62.35 & 72.77 & 10.63 & 30.96M &  5.54 \\
& \textbf{DCSNet(Ours)}  & \textbf{77.43} & \textbf{64.48} & 76.87 & \textbf{9.90} & 59.42M &  46.86 \\
\bottomrule
\end{tabular}%
} 
\end{table}

\textbf{Qualitative Results.}
    To complement the quantitative metrics with an intuitive morphological evaluation, we visualize the segmentation results of DCSNet against representative baselines across all three datasets.
    
\begin{figure}[htb]
    \centering
    \includegraphics[width=1.0\linewidth]{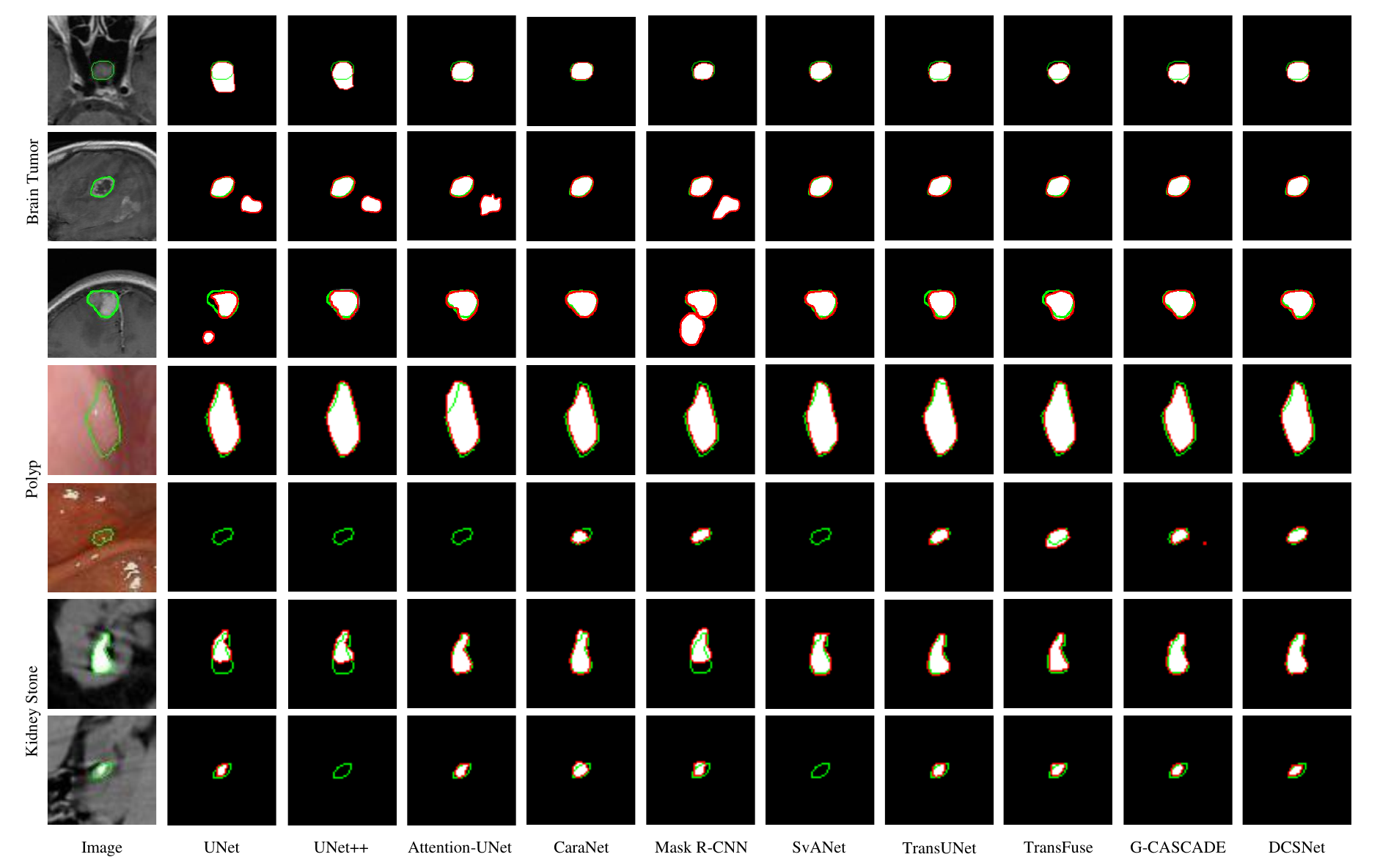}
    \caption{Zoomed-in visualization of the segmentation results with highlighted boundaries. The contours explicitly demonstrate the superior capability of our DCSNet in preserving detailed boundary structures and tightly adhering to the highly irregular shapes of small lesions, overcoming the over-segmentation or blocky artifacts observed in other baselines.}
    \label{fig:zoom_in}
\end{figure}

    As illustrated in Fig.~\ref{fig:zoom_in}, zoomed-in contour overlays are utilized to explicitly assess target localization and boundary adherence. Classical CNN architectures (e.g., UNet, Attention-UNet) exhibit a severe vulnerability to under-segmentation and complete localization failure for micro-lesions, particularly in low-contrast scenarios like the Kidney Stone dataset. Conversely, global context-driven methods (e.g., TransUNet, TransFuse) struggle to suppress false positives, directly resulting in the over-segmentation of surrounding background tissue. DCSNet substantially mitigates these visual artifacts, which are fundamentally driven by extreme class imbalance. By employing the DGHC module to constrain feature extraction strictly within object-centric regions, the framework effectively suppresses background noise and facilitates the robust localization of highly sparse targets.
    
    Beyond spatial localization, small medical lesions exhibit inherent boundary complexity. As observed in the detailed contour overlays, baseline methods, including the recent G-CASCADE, produce overly smooth, blocky, or disjointed boundaries that fail to capture the true anatomical morphology. In contrast, the contours predicted by DCSNet are noticeably sharper and align more tightly with the ground-truth delineations. This visual fidelity directly corroborates our superior 95HD metrics and highlights the structural efficacy of the MSFA module. By dynamically fusing multiscale cropped features and amplifying local spatial cues, DCSNet alleviates the resolution degradation typical of deep networks, yielding the highly precise boundary recovery desired for clinical diagnostics.

\begin{table}[htbp]
\centering
\caption{Ablation study on the Brain Tumor dataset. The effectiveness of the Detection-guided Hierarchical Cropping (DGHC) and Multiscale Feature Aggregation (MSFA) modules is evaluated. `\checkmark' denotes the inclusion of the respective module.}
\label{tab:ablation}
\resizebox{\textwidth}{!}{%
\begin{tabular}{cccccccc}
\toprule
\multicolumn{2}{c}{Modules} & \multirow{2}{*}{Dice $\uparrow$} & \multirow{2}{*}{IoU $\uparrow$} & \multirow{2}{*}{Precision $\uparrow$} & \multirow{2}{*}{95HD $\downarrow$} & \multirow{2}{*}{Params} & \multirow{2}{*}{GFLOPs} \\
\cmidrule(lr){1-2}
DGHC & MSFA & & & & & & \\
\midrule
 & & 74.91 & 64.07 & 78.92 & 25.07 & 55.46M & 77.89 \\
\checkmark & & 79.68 & 69.07 & 81.28 & 24.21 & 55.46M & 77.89 \\
 & \checkmark & 80.06 & 68.74 & 82.99 & 16.02 & 59.42M & 83.32 \\
\checkmark & \checkmark & \textbf{83.67} & \textbf{74.04} & \textbf{85.24} & \textbf{12.18} & 59.42M & 83.32 \\
\bottomrule
\end{tabular}%
} 
\end{table}

\subsection{Ablation Study}
    To validate the individual architectural contributions of the proposed DCSNet, a comprehensive ablation study is conducted on the Brain Tumor dataset. The baseline model is constructed by completely bypassing the DGHC and MSFA modules. We incrementally integrate these components to quantitatively evaluate their impact on segmentation performance, computational complexity, and parameter overhead. The ablation results are detailed in Table \ref{tab:ablation}.

\textbf{Effectiveness of the DGHC Module.} 
    As demonstrated in the second row of Table \ref{tab:ablation}, integrating the DGHC module into the baseline yields substantial accuracy improvements without introducing any additional parameters (remaining at 55.46M). The Dice score increases from 74.91\% to 79.68\%, and the IoU sees a commensurate 5.0\% boost. This enhanced accuracy incurs no computational penalty, with the overhead remaining virtually identical at 77.89 GFLOPs. This strictly cost-free performance gain confirms that utilizing region proposals to constrain segmentation attention effectively eliminates false positives and enhances target localization, all while maintaining absolute computational efficiency.

\textbf{Effectiveness of the MSFA Module.} 
    Deploying the MSFA module independently triggers a critical reduction in boundary errors. Compared to the baseline, the addition of MSFA drives the 95HD metric down significantly from 25.07 to 16.02, while simultaneously elevating the Dice score to 80.06\%. While this integration incurs a modest parameter increase (+3.96M) and slightly higher computational cost (+5.43 GFLOPs) due to the multiscale aggregation mechanism, the drastic improvement in boundary fidelity justifies the architectural trade-off. It proves that MSFA successfully refines local spatial details and adheres more tightly to irregular lesion contours.

\textbf{Synergy of DGHC and MSFA.} 
    The complete DCSNet achieves optimal performance across all metrics, yielding a peak Dice score of 83.67\% and minimizing the 95HD to 12.18. These quantitative improvements reflect a meticulously designed architectural sequence rather than a simple superposition of isolated components. Functionally, DGHC first tackles extreme class imbalance by filtering out expansive background noise to secure robust target localization. To resolve the subsequent challenge of inherent boundary complexity, the MSFA module is deployed, explicitly tailored to process these localized representations. By restricting the receptive field to object-centric areas, DGHC shields the self attention mechanism from background dilution, empowering MSFA to concentrate its full representational capacity on delineating intricate contours. Operating on these refined spatial cues, the pixel-adaptive fusion substantially alleviates boundary ambiguity, facilitating highly precise micro-lesion segmentation.

\begin{table}[htbp]
\centering
\caption{Ablation study on the base RoI scale ($H$) using the colorectal polyp dataset. The feature pyramid resolutions are determined by the base scale $H$ as $\{H, H/2, H/4, H/8\}$. The optimal scale is highlighted in bold.}
\label{tab:roi_scale}
\resizebox{0.9\textwidth}{!}{%
\begin{tabular}{cccccc}
\toprule
Base Scale ($H$) & Feature Pyramid Resolution & Dice $\uparrow$ & IoU $\uparrow$ & Precision $\uparrow$ & 95HD $\downarrow$ \\
\midrule
128 & $128, 64, 32, 16$ & 87.49 & 77.36 & 83.72 & 4.67 \\
64  & $64, 32, 16, 8$   & \textbf{90.34} & \textbf{82.59} & \textbf{88.14} & \textbf{3.23} \\
32  & $32, 16, 8, 4$    & 88.04 & 78.73 & 81.26 & 4.83 \\
\bottomrule
\end{tabular}%
}
\end{table}

    To determine the optimal cropping resolution for the DGHC module, we conduct an additional ablation study on the base RoI scale $H$. Rather than scaling the cropped regions proportionally to the varying input resolutions of different datasets, we map all multiscale features to a fixed-resolution pyramid to maintain a standardized semantic receptive field for the subsequent Transformer encoder. We evaluate three base scales using the colorectal polyp dataset: $H=128$, $H=64$, and $H=32$. 
    
    As summarized in Table \ref{tab:roi_scale}, setting the base scale to $H=64$ yields the optimal performance across all metrics. A smaller base scale ($H=32$) results in a noticeable performance drop, with the 95HD metric deteriorating from 3.23 to 4.83. This indicates that an excessively low resolution irreversibly destroys the fine-grained structural details and boundary topological information of the micro-lesions. Conversely, increasing the base scale to $H=128$ fails to provide further performance gains and instead yields suboptimal metrics. This phenomenon occurs because an excessively large cropping region inevitably introduces redundant background noise, diluting the foreground object features and exacerbating the class imbalance problem within the RoI. Consequently, we empirically fix the base scale at $H=64$ across all experiments to ensure optimal boundary preservation and robust generalization.
    
\section{Conclusion}
    In this paper, we propose DCSNet to overcome the inherent challenges of extreme class imbalance and inherent boundary complexity in small medical object segmentation. By synergizing detection-guided cropping with multiscale feature aggregation, our framework transforms standard global prediction into a highly focused, localized refinement process, which effectively mitigates scale variations and alleviates boundary ambiguity. Comprehensive experiments across three diverse medical datasets, including brain tumor MRI, colorectal polyp endoscopy, and kidney stone CT, validate the effectiveness of DCSNet. It yields highly competitive performance in regional overlap metrics such as Dice and IoU, while achieving significant improvements in boundary precision measured by 95HD. These results demonstrate its reliable capability to delineate the highly irregular contours of small lesions without falling into the trap of over-smoothing. Furthermore, this coarse-to-fine localization paradigm provides an adaptable baseline for the broader pattern recognition community to segment extremely sparse targets in other noisy or low-contrast environments.

    While DCSNet demonstrates robust performance, there remains room for further optimization. The integration of multistage feature extraction and Transformer-based aggregation introduces a moderate computational overhead. Moreover, the current 2D formulation processes slices independently, which does not fully exploit the inter-slice topological context inherent in spatial scans. Moving forward, our future work will focus on exploring lightweight structural adaptations to enhance computational efficiency, facilitating real-time clinical deployment. Furthermore, we plan to extend this detection-guided refinement paradigm to 3D volumetric data, allowing the model to fully capture spatial continuity without sacrificing boundary precision.

\section*{Acknowledgments}
 This work was supported by the National Natural Science Foundation of China under Grant No. 62495064, the Chengdu City's Collaborative Innovation Project for Industrial Chain under Grant No. 2026-XT00-00018-GX, and the Natural Science Foundation of Sichuan Province under Grant No. 2026NSFSC1488.   
\bibliography{reference}

\end{document}